\documentclass{article}
\usepackage{ijcai17}

\usepackage{times}
\usepackage{url}
\usepackage{amsmath}
\usepackage{amsthm}
\usepackage{amsfonts}
\usepackage{xspace}
\usepackage{multirow}
\usepackage{algorithmicx}
\usepackage[algo2e]{algorithm2e}

\usepackage{algpseudocode}
\usepackage{algorithm}
\usepackage{graphicx}
\usepackage{caption,subcaption}

\DeclareMathOperator*{\argmax}{arg\,max}

\newcommand{\lif}[0]{
  \leftarrow
}

\newcommand{\vt}[1]{
\mathbf{#1}
}
\newcommand{\fn}[1]{
{\it #1}
}

\newcommand{\fzand}[0]{
  \wedge
}
\newcommand{\bAnd}[0]{
  \bigwedge
}
\newcommand{\fzor}[0]{
  \vee
}
\newcommand{\bOr}[0]{
  \bigvee
}

\newcommand{\fziff}[0]{
  \leftrightarrow
}

\newcommand{\pr}[1]{
  \mathrm{#1}
}
\newcommand{\En}[0]{ 
  \fn{E}
}

\newtheorem{example}{Example}

\title{Unsupervised Neural-Symbolic Integration}
\author{Son N. Tran\\ 
The Australian E-Health Research Center, CSIRO  \\
son.tran@csiro.au
}

\begin{document}
\maketitle
\begin{abstract}
Symbolic has been long considered as a language of human intelligence
while neural networks have advantages of robust computation and
dealing with noisy data. The integration of neural-symbolic can offer
better learning and reasoning while providing a means for
interpretability through the representation of symbolic
knowledge. Although previous works focus intensively on supervised
feedforward neural networks, little has been done for the unsupervised
counterparts. In this paper we show how to integrate symbolic
knowledge into unsupervised neural networks. We exemplify our approach
with knowledge in different forms, including propositional logic for
DNA promoter prediction and first-order logic for understanding family
relationship.
\end{abstract}

\section{Introduction}
An interesting topic in AI is integration of symbolic and neural
networks, two different information processing paradigms. While the
former is the key of higher level of intelligence the latter is well
known for a capability of effective learning from data. In the last two decades,
researchers have been working on the idea that combination of the two
should offer joint
benefits \cite{Towel_1994,Smolensky_1995,Garcez_1999,Valiant_2006,Garcez_2008,Leo_2011,Franca_2014,Son_2016}.

In previous work, supervised neural networks have been used
intensively for the integration based on the analogy of {\it modus
ponens} inference with symbolic rules and forward passing in neural
networks \cite{Towel_1994,Garcez_1999}. In such networks, due to the
discriminative structures only a subset of variables can be inferred,
i.e. the variables in the left hand of {\it if-then} $\lif$
formulas. This may limit their use in general reasoning. Unsupervised
network, on the other hand, offers more flexible inference mechanism
which seems more suitable for symbolic reasoning. Let us consider an
XOR example $\pr{z} \fziff (\pr{x} \oplus \pr{y})$. Here, given the
truth values of any two variables one can infer the rest. For
supervised networks, a class variable must be discriminated from the
others and only it can be inferred. An unsupervised network, in
contrast, do not require such discrimination.

Encoding symbolic knowledge in an unsupervised neural network needs a
mechanism to convert symbolic formulas to the network without loss of
generality. In previous work, Penalty logic shows that any
propositional formula can be represented in a symmetric connectionist
network (SCN) where inference with rules is equivalent to minimising the
network's energy. However, SCN uses dense connections of hidden and
visible units which make the inference very computational. Recent work
shows that any propositional formula can be represented in restricted
Boltzmann machines \cite{Son_2017}. Different from Penalty logic, here the RBM
is a simplified version of SCN where there is no visible-visible and
hidden-hidden connections. This makes inference in RBMs is easier.

Several attempts have been made recently to integrate symbolic
representation and RBMs \cite{Leo_2011,Son_2016}. Despite achieving
good practical results they are still heuristic. In this
paper, we show how to encode symbolic knowledge in both propositional
and first-order forms into the RBM by extending the theory
in \cite{Son_2017}.

The remainder of this paper is organized as
follows. Section \ref{sec:cf} review the idea of Confidence rule, a
knowledge form to represent symbolic formulas in
RBMs. In section \ref{sec:ke} we show how to encode
knowledge into RBMs. Section \ref{sec:ee} presents the empirical
verification of our encoding approach and Section \ref{sec:d}
concludes the work.
\section{Confidence Rules: Revisit}
\label{sec:cf}

A confidence rule \cite{Son_2013,Son_2016} is a propositional formula in the form:
\begin{equation}
c:\pr{h} \fziff \bAnd_t \pr{x}_t \fzand \bAnd_k \neg \pr{x}_k
\end{equation}
where $\pr{h}$ is called {\it hypothesis}, $c$ is a non-negative real
value called {\it confidence value}.  Inference with a confidence rule
is to find the model that makes the hypothesis $\pr{h}$ holds. If
there exist a target variable $y$ the inference of such variable will be similar to
{\it modus ponens}, as shown in Table \ref{tab:modus_ponens}
\begin{table}[ht]
\centering
\begin{tabular}{l l}
Confidence rule inference  &  Modus ponens\\
{\scriptsize $\pr{h} \fziff \bAnd_{t \in T} \pr{x}_{t} \fzand \bAnd_{k \in K}  \neg \pr{x}_k \fzand \pr{y}$} &{\scriptsize $\pr{y} \lif \bAnd_t \pr{x}_t \fzand \bAnd_k \neg \pr{x}_k$}\\
{\scriptsize $\{\pr{x}_t,\neg \pr{x}_k| \text{ for } \forall t \in T, \forall k \in K\}$} & {\scriptsize$\{\pr{x}_t,\neg \pr{x}_k| \text{ for } \forall t \in T, \forall k \in K\}$}\\
\line(1,0){100} & \line(1,0){100} \\
{\scriptsize $\pr{y}$}& {\scriptsize$\pr{y}$}
\end{tabular}

\caption{Confidence rule and Modus ponens}
\label{tab:modus_ponens}
\end{table}

An interesting feature of Confidence rules is that one can represent
them in an RBM where Gibbs sampling can be seen equivalently as
maximising the total (weighted) satisfiability \cite{Son_2017}. If a
knowledge base is converted into Confidence rules then we can take the
advantage of the computation mechanism in such neural networks for
efficient inference. The equivalence between confidence rules and an
RBM is defined in that the satisfiability of a formula is inversely
proportional to the energy of a network:

\begin{equation*}
s_\varphi(\vt{x}) = -a\En_{rank}(\vt{x}) + b
\end{equation*}

where $s_\varphi$ is the truth value of the formula $\varphi$ given an
 assignment $\vt{x}$; $\En_{rank}(\vt{x}) = min_\vt{h}\En(\vt{x},\vt{h})$ is
 the energy function minimised over all hidden variables; $a>0$,$b$ are scalars.
 
By using disjunctive normal form (DNF) to present knowledge Confidence
rules attract some criticism for practicality since it is more popular
to convert a formula to a conjunctive normal form of polynomial
size. However we will show that Confidence rules are still very useful
in practice. In fact, in such tasks as knowledge extraction, transfer,
and integration Confidence rules have been already
employed \cite{Leo_2011,Son_2013,Son_2016}. For knowledge integration previous
work separates the {\it if-and-only-if} symbol in Confidence rules into
two {\it if-then} rules to encode in a hierarchical
networks \cite{Son_2016}. In this work, we show that such separation is not
necessary since any propositional {\it if-then} formulas can be
efficiently converted to Confidence rules. The
details are in the next section.
\section{Knowledge Encoding}
\label{sec:ke}
In many cases background knowledge presents a set of {\it if-then}
formulas (or equivalent Horn clauses). This section shows how to
convert them into Confidence rules for both propositional and
first-order logic forms
\subsection{Proposition Logic}
\label{subsec:ke_prop}
A propositional {\it if-then} formula has the form
\begin{equation*}
c:\pr{y} \lif \bAnd_t \pr{x}_t \fzand \bAnd_k \neg \pr{x}_k
\end{equation*}
which can be transformed to a DNF as:
\begin{equation*}
c:(\pr{y} \fzand \bAnd_t \pr{x}_t \fzand \bAnd_k \neg \pr{x}_k) \fzor \bOr_t (\neg \pr{x}_t) \fzor \bOr_k (\pr{x}_k)
\end{equation*}
and then to the confidence rules:
\begin{equation*}
\begin{aligned}
&c: \pr{h}_y \fziff \pr{y} \fzand \bAnd_t \pr{x}_t \fzand \bAnd_k \neg \pr{x}_k \\
&c: \pr{h}_t \fziff \neg \pr{x}_t \text{ for } \forall t \\
&c: \pr{h}_k \fziff \pr{x}_k \text{ for } \forall k \\ 
\end{aligned}
\end{equation*}

Encoding these rules into an RBM does not guarantee the
equivalence. This is because it violates the condition that the DNF of
a formula should {\it have at most one conjunct is true given an
assignment} \cite{Son_2017}. Fortunately this can be solved by grouping
$\neg \pr{x}_t$, $\pr{x}_k$ with a max-pooling hidden unit which
results in an RBM with the energy function as{\footnote{The proof is
similar as in \cite{Son_2017}}\footnote{$0 <\epsilon <1$}:
\begin{equation}
\begin{aligned}
\En = &- c\times h_y(y + \sum_t x_t - \sum_k x_k - |T|-1+\epsilon)\\
      &- c\times h_p\max(\{-x_t+\epsilon,x_k-1+\epsilon|t\in T, k\in K\})
\end{aligned}
\end{equation}

Here a max pooling hidden unit represents a hypothesis:
$\pr{h}_p \fziff \bOr_t \pr{h}_t \fzor \bOr_k \pr{h}_k$ which, in this
case, can be written as:
$c:\pr{h}_p \fziff \bOr_t \neg \pr{x}_t \fzor \bOr_k \pr{x}_k$. The
final set rules are:
\begin{equation*}
\begin{aligned}
&c: \pr{h}_y \fziff \pr{y} \fzand \bAnd_t \pr{x}_t \fzand \bAnd_k \neg \pr{x}_k \\
&c:\pr{h}_p \fziff \bOr_t \neg \pr{x}_t \fzor \bOr_k \pr{x}_k \\
\end{aligned}
\end{equation*}

\begin{example} Let us consider the formula: 
$5: \pr{y} \lif \pr{x}_1 \fzand \neg \pr{x}_2$ which would be
converted to DNF as: $5:
(\pr{y} \fzand \pr{x}_1 \fzand \neg \pr{x}_2) \fzor
(\neg \pr{x}_1) \fzor (\pr{x}_2)$, and then to an RBM with the energy
function: $\En = -5h_1(y + x_1 - x_2 -1.5) -
5h_2\max(-x_1+0.5,x_2-0.5)$. Table \ref{fig:rl_rbm} shows the equivalence between
the RBM and the formula.

\begin{table}[ht]
\centering
\begin{tabular}{|c|c|c|c|c|}
\hline
$x_1$ &$x_2$ &$y$ & $s_\varphi$ & $\En_{rank}$\\
\hline
0 & 0 & 0 & 1 & -2.5 \\
0 & 0 & 1 & 1 & -2.5 \\
0 & 1 & 0 & 1 & -2.5 \\
0 & 1 & 1 & 1 & -2.5 \\
1 & 0 & 0 & 0 & 0    \\
1 & 0 & 1 & 1 & -2.5 \\
1 & 1 & 0 & 1 & -2.5 \\
1 & 1 & 1 & 1 & -2.5 \\
\hline
\end{tabular}
\caption{Energy of RBM and truth values of the formula $5: \pr{y} \lif \pr{x}_1 \fzand \neg \pr{x}_2$}
\label{fig:rl_rbm}
\end{table}
where $s_\varphi$ is (unweighted) truth values of the formula. This
indicates the equivalence between the RBM and the formula as:
\begin{equation*}
s_\varphi(\pr{x1},\pr{x2},\pr{y}) = - \frac{1}{2.5} \pr\En_{rank}(x,y,z) + 0
\end{equation*}
\end{example}
\subsection{First-order Logic}
\label{subsec:ke_fo}
A first order logic formula can also be converted into a set of
Confidence rules. First, let us consider a predicate: $P(x,y)$ which
one can present in a propositional DNF as:
\begin{equation*}
\bOr_{a,b|P(a,b)=true} \pr{p}_{x=a} \fzand \pr{p}_{y=b} \fzand \pr{p}_P
\end{equation*}

where $(a,b)$ are the models of $P(x,y)$; $\pr{p}_{x=a}$ and
$\pr{p}_{y=b}$ are the propositions that are $true$ if $x=a$ and $y=b$
respectively, otherwise they are $false$; $\pr{p}_P$ is the proposition
indicating if the  value of $P(a,b)$. Each conjunct in this DNF
then can be represented as a Confidence rule.

Now, let us consider a first-order formula which we are also able to
present in a set of Confidence rules. For example, a clause
$\varphi$ as:
\begin{equation*}
\forall_{x,y,z} \text{son}(x,z) \lif \text{brother}(x,y) \fzand \text{has\_father}(y,z)
\end{equation*}

can be converted into:
\begin{equation*}
\begin{aligned}
\bOr_{a,b,c|\varphi=true} & (\pr{p}_{x=a} \fzand \pr{p}_{y=b}\fzand \pr{p}_{z=c} \fzand \pr{p}_\text{son} \fzand \pr{p}_\text{brother} \fzand \pr{p}_\text{has\_father})\\
& \fzor (\neg \pr{p}_{x=a} \fzor \neg \pr{p}_{y=b} \fzor \neg \pr{p}_{z=c} \fzor \neg \pr{p}_\text{brother} \fzor \neg \pr{p}_\text{has\_father})
\end{aligned}
\end{equation*}


If one want to encode the background knowledge through its
samples, for example:
\begin{equation*}
\begin{aligned}
\text{son}(James,Andrew) & \lif \text{brother}(James,Jen)\\
&\fzand \text{has\_father}(Jen,Andrew)
\end{aligned}
\end{equation*}
then we can convert it into confidence rules:
\begin{equation*}
\begin{aligned}
&c:\pr{h}_1 \fziff \pr{james} \fzand \pr{jen} \fzand \pr{andrew} \fzand \pr{\textbf{son}} \fzand \pr{\textbf{brother}} \fzand \pr{\textbf{has\_father}}\\
&c: \pr{h}_p \fziff \neg \pr{james} \fzor \neg \pr{jen} \fzor \neg \pr{andrew} \fzor \neg \pr{\textbf{brother}} \fzor \neg \pr{\textbf{has\_father}}
\end{aligned}
\end{equation*}
In practice, in many cases we are only interested in inferring the
predicates therefore we can omit $\neg \pr{james}$, $\neg \pr{jen}$,
$\neg \pr{andrew}$ from the second rule.

\section{Empirical Evaluation}
\label{sec:ee}
In this section we apply the encoding approaches discussed in the
previous section to integrate knowledge into unsupervised networks.
\subsection{DNA promoter}
The DNA promoter dataset consist of a background theory with 14
logical {\it if-then} rules \cite{Towel_1994}. The rules includes four symbols
$contact$, $minus_{10}$, $minus_{35}$, $conformation$ which are not
observed in the data. This is suitable for hierarchical models as shown
in previous works \cite{Towel_1994,Son_2016}. In this experiment we
group the rules using {\it hypothetical syllogism} to eliminate the
unseen symbols. After that we encode the rules in an RBM following
the theory in Section \ref{subsec:ke_prop}. The confidence values are selected
empirically.

We test the normal RBMs and the RBMs with encoded rule using
leave-one-out method, both achieve $100\%$ accuracy. In order to
evaluate the effectiveness of our approach we partition the data into
nine different training-test sets with number of training samples are
$10$, $20$, $30$, $40$, $50$, $60$, $70$, $80$, $90$. All experiments
are repeated 50 times and the average results are reported in
Figure \ref{fig:rbm_dna_exp}. We perform the prediction using both
Gibbs sampling and conditional distribution $P(y|\vt{x})$. In
particular, Figure \ref{fig:dna_gibb} shows the prediction results
using 1-step Gibbs sampling where the input is fixed to infer the
hidden states and then to infer the label unit. In
Figure \ref{fig:dna_cond} the results show the prediction accuracy
achieved by inferring the label unit from the conditional
distribution. As we can see, in both cases the integrated RBMs perform
better than the normal RBMs on small training sets with number of
training sample is less than 60. With larger training sets, the rules
are no longer effective since the training samples are adequate to
generalise the model to achieve $100\%$ accuracy. 
\begin{figure}[ht]
\centering
\begin{subfigure}{0.23\textwidth}
\includegraphics[width=1.1\textwidth]{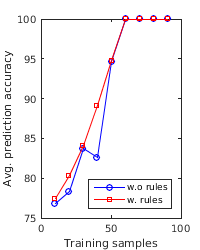}
\caption{Infer with Gibbs sampling}
\label{fig:dna_gibb}
\end{subfigure}
\begin{subfigure}{0.23\textwidth}
\includegraphics[width=1.1\textwidth]{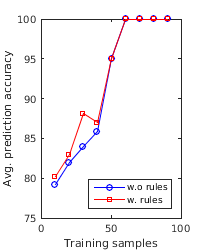}
\caption{Infer with $P(y|\vt{x})$}
\label{fig:dna_cond}
\end{subfigure}
\caption{RBMs without rules v.s RBMs with rules}
\label{fig:rbm_dna_exp}
\end{figure}

\subsection{Kinship}
In this experiment, we use the approach discussed in
Section \ref{subsec:ke_fo} for relation discovery and reasoning tasks
with Kinship dataset \cite{Hinton_1986b,Ilya_2008}. Here given a set
of examples about relations we perform two type of reasoning: (1) what
is relation between two people, i.e $?(x,y)$; and (2) a person has a
relation $R$ with whom, i.e. $R(x,?)$.  Previous approaches are using
matrices/tensors to represents the relations making it difficult to
explain \cite{Ilya_2008}. In this work, since only predicates are
given, we encode the examples for the predicates in an unsupervised
network as shown earlier in Section \ref{subsec:ke_fo}. This
constructs the left part of the integrated model in
Figure \ref{fig:kinship_model}. In the right part, we model the
unknown clauses by using a set of hidden units. The idea here is that
by inferring the predicates using the encoded rules in the left part
we can capture the relationship information, from which the desired
relation is inferred by reconstruction of such relationship in the
right part. In this experiment, we use auto-encoder \cite{Bengio_2009}
for the right part for the purpose of efficient learning. The whole
process is described in Algorithm \ref{al1}.

\begin{figure}[ht]
\includegraphics[width=0.49\textwidth]{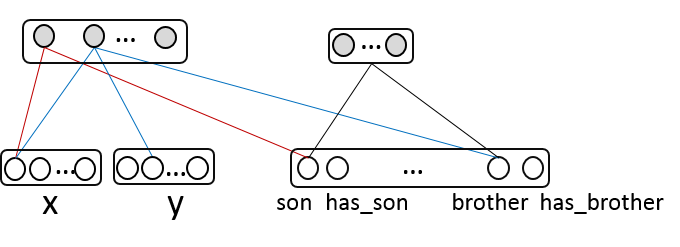}
\caption{Encoding Kinship examples}
\label{fig:kinship_model}
\end{figure}

Let us take an example where one wants to find the relation between
two people $R(Marco, Pierro) = ?$. First, we use the other examples to
construct an integrated model as in
Figure \ref{fig:kinship_model}. After that, we train the auto-encoder
in the right part using unsupervised learning algorithm, then we
extract the relation features from Marco and Pierro as shown in
Table \ref{tab:marco_pierro}. In that table we also show the reconstructed scores for
all the relations where $\pr{son}$ is the correct one.
\begin{table*}[ht]
\centering
\begin{tabular}{l l l}
Marco, Pierro & 1.000: has\_father & Marco has father who is Pierro \\
\hline
\multirow{ 9}{*}{Marco} & 0.191:has\_wife           & Marco has wife\\
                              & 0.191:husband       & Marco is a husband of someone\\
                              & 0.191:has\_mother   & Marco has  mother\\
                              & 0.191:father        & Marco is a father of someone\\              
                              & 0.191:has\_daughter & Marco has daughter\\
                              & 0.191:son           & Marco has is a son of someone\\
                              & 0.191:has\_son      & Marco has son\\
                              & 0.191:has\_sister   & Marco has sister\\
                              & 0.191:brother       & Marco is a brother of someone\\
\hline                              
\multirow{ 4}{*}{Pierro}     & 0.191: has\_wife     & Pierro has  wife\\
                             & 0.191: husband       & Pierro is a husband of someone\\ 
                             & 0.191: father        & Pierro is a father of someone\\
                             & 0.191: has\_daughter & Pierro has  daughter\\
\hline
\multirow{ 2}{*}{Reconstruct (possible relations)} & \multicolumn{2}{l}{0.008:wife 0.008:husband 0.001:mother 0.016:father 0.045:daughter \textbf{0.252:son}}\\
                                  &\multicolumn{2}{l}{0.005:sister 0.013:brother 0.001:aunt 0.002:uncle 0.003:niece 0.002:nephew}
\end{tabular}
\caption{Relation features}
\label{tab:marco_pierro}
\end{table*}

\begin{algorithm}[h]
\KwData{Examples: E, Question: R(a,b)}
 \KwResult{R} \hskip .2cm Encode all examples in an RBM: N\\
  Initialise $\mathcal{D}=\emptyset$\\ \For{each example $R(x,y)$ in
  E}{ $f$ = INFER(N,x,y)\\ Add $f$ to $\mathcal{D}$\\ } \hskip .2cm
  Train an Auto-Encoder (AE) on $\mathcal{D}$\\ $f$ = INFER(N,a,b)\\
  Reconstruct $\hat{f}$ using AE\\ Return R =
  $\argmax_R(\hat{f}_R)$\Comment{Return the unseen relation where the
  reconstruction feature have the highest value.}
\caption{}
\label{al1}
\end{algorithm}

\begin{algorithm}[ht]
    \begin{algorithmic}[1]
        \Function{INFER}{$N,a,b$}
         \State Infer direct relation between a,b\;
         \State Infer possible relations of a:  R(a,*)\;
         \State Infer possible relations of b: R(*,b)\;
         \State f = concatenation of all relations\;
         \State Return f
       \EndFunction
  \end{algorithmic}
\end{algorithm}

We test the model on answering the question $R(x,y)=?$ using
leave-one-out validation which achieve $100\%$ accuracy. We also use
the integrated model to reason about whom one has a relation
with. This question may have more than one answer, for example
$\text{son}(Athur,?)$ can be either $Cristopher$ or $Penelope$. We
randomly select 10 examples for testing and repeat it for $5$
times. If the designate answers are in the top relations with highest
reconstructed features then we consider this as correct, otherwise we
set it as wrong. The average error of this test is $0\%$. However,
when we increase the number of test samples to $20$ and $30$ the
average errors grow to $2.8\%$ and $6.8\%$ respectively. For
comparison, the matrices based approach such as \cite{Ilya_2008}
achieves $0.4\%$,$1.2\%$,$2.0\%$ average error rates for $10$, $20$,
$30$ test examples respectively. Note that, such approach and many
others \cite{Socher_2013} model each relation by a matrix/tensor while
in this experiment we share the parameters across all relations. Also,
the others use discriminative learning while we use unsupervised
learning. The purpose of this is to exemplify the encoding technique
we proposed earlier in this paper. Improvement can be achieved if
similar methods are employed.
\section{Conclusions}
\label{sec:d}
The paper shows how to integrate symbolic knowledge into unsupervised
neural networks. This work bases on the theoretical finding that any
propositional formula can be represented in RBMs \cite{Son_2017}. We
show that converting background knowledge in the form of {\it if-then}
rules to Confidence rules for encoding is efficient. In the
experiments, we evaluate our approaches for DNA promoter prediction
and relationship reasoning to show the validity of the approach. 
\bibliographystyle{named}
\bibliography{../IJCAI_17/compstat,../IJCAI_17/deepnet,../IJCAI_17/logprob,../IJCAI_17/other,../IJCAI_17/topicmodel,../IJCAI_17/connectionist,../IJCAI_17/ijcai17,../IJCAI_17/nlp,../IJCAI_17/reinforcement,../IJCAI_17/statml,../IJCAI_17/transflearning}

\end{document}